\documentclass[journal]{IEEEtran}
\usepackage[switch]{lineno}
\newcommand*\patchAmsMathEnvironmentForLineno[1]{
  \expandafter\let\csname old#1\expandafter\endcsname\csname #1\endcsname
  \expandafter\let\csname oldend#1\expandafter\endcsname\csname end#1\endcsname
  \renewenvironment{#1}
     {\linenomath\csname old#1\endcsname}
     {\csname oldend#1\endcsname\endlinenomath}}
\newcommand*\patchBothAmsMathEnvironmentsForLineno[1]{
  \patchAmsMathEnvironmentForLineno{#1}
  \patchAmsMathEnvironmentForLineno{#1*}}
\AtBeginDocument{
\patchBothAmsMathEnvironmentsForLineno{equation}
\patchBothAmsMathEnvironmentsForLineno{align}
\patchBothAmsMathEnvironmentsForLineno{flalign}
\patchBothAmsMathEnvironmentsForLineno{alignat}
\patchBothAmsMathEnvironmentsForLineno{gather}
\patchBothAmsMathEnvironmentsForLineno{multline}
}

\usepackage{cite}                      
\usepackage{tabu}                      
\usepackage{booktabs}                  
\usepackage{subfig}
\usepackage{times}
\usepackage{epsfig}
\usepackage{graphicx}
\usepackage{amsmath}
\usepackage{amssymb}
\usepackage{algorithm}
\usepackage{color}
\usepackage{bbm}
\usepackage{bm}
\usepackage{multirow}
\usepackage{comment}
\usepackage{here}
\usepackage{booktabs}
\usepackage{arydshln}
\usepackage{hyperref}
\usepackage{float}
\usepackage[noend]{algpseudocode} 
\usepackage{algorithmicx}
\usepackage{subcaption}
\usepackage[utf8]{inputenc}

\usepackage[dvipsnames]{xcolor}

\ifCLASSINFOpdf
\else
\fi
\begin{document}
%
\title{3D Human-Human Interaction Anomaly Detection}
%
%

\author{Shun~Maeda*,
        Chunzhi~Gu*,
        Koichiro~Kamide, 
        Katsuya Hotta,
        Shangce~Gao,
        and~Chao~Zhang
\thanks{S. Maeda* is with the School of Engineering, University of Fukui, Fukui, Japan (msd24006@u-fukui.ac.jp).}
\thanks{C. Gu* is with the Faculty of Engineering, University of Fukui, Fukui, Japan (czgu@ieee.org).}
\thanks{K. Hotta is with the Faculty of Engineering, Iwate University, Iwate, Japan (hotta@iwate-u.ac.jp).}
\thanks{K. Kamide, S. Gao and C. Zhang are with the Department of Engineering, University of Toyama, Toyama, Japan (m25c1014@ems.u-toyama.ac.jp, gaosc@eng.u-toyama.ac.jp, zhang@eng.u-toyama.ac.jp).}
\thanks{*Equal contribution.}
\thanks{Corresponding author: C. Zhang.}
}
\maketitle

\begin{abstract}
Human-centric anomaly detection (AD) has been primarily studied to specify anomalous behaviors in a single person. However, as humans by nature tend to act in a collaborative manner, behavioral anomalies can also arise from human-human interactions. Detecting such anomalies using existing single-person AD models is prone to low accuracy, as these approaches are typically not designed to capture the complex and asymmetric dynamics of interactions. In this paper, we introduce a novel task, Human-Human Interaction Anomaly Detection (H2IAD), which aims to identify anomalous interactive behaviors within collaborative 3D human actions. To address H2IAD, we then propose Interaction Anomaly Detection Network (IADNet), which is formalized with a Temporal Attention Sharing Module (TASM). Specifically, in designing TASM, 
we share the encoded motion embeddings across both people such that collaborative motion correlations can be effectively synchronized. Moreover, we notice that in addition to temporal dynamics, human interactions are also characterized by spatial configurations between two people. We thus introduce a Distance-Based Relational Encoding Module (DREM) to better reflect social cues in H2IAD. The normalizing flow is eventually employed for anomaly scoring. Extensive experiments on human-human motion benchmarks demonstrate that IADNet outperforms existing Human-centric AD baselines in H2IAD.

\end{abstract}


%
\IEEEpeerreviewmaketitle

\section{Introduction}

\IEEEPARstart{T}{he} detection of anomalous human behaviors has long been an important topic in computer vision due to its wide range of applications, from surveillance systems \cite{diehl2002real, sultani2018real} to human-object interactions \cite{wang2025hoi2anomaly}. In recent years, increasing efforts \cite{doshi2020continual, tao2024feature, rodrigues2020multi, rai2024video, georgescu2021anomaly, chen2023tevad} have been made to handle anomalies within human activities. 

Conventional techniques \cite{flaborea2023multimodal,gao2024multi,hirschorn2023normalizing,karami2025graph} mostly focus on handling anomalous human behaviors in videos. Typically, these methods learn from videos of normal human activities in an unsupervised fashion to identify whether a given new human recording includes anomalies. 
Despite the generality, this line of research aims to recognize the abnormality from general 2D human events, and does not learn the underlying semantics from human motions. To better model human actions to realize AD at the semantic level, one recent work \cite{maeda2025frequency} proposed the task of human action AD, which focuses on detecting anomalous action patterns by defining the exact semantic action type to constitute the normal data. It also developed a multi-level framework to fuse global and local semantic cues to distinguish subtle action anomalies. Overall, existing human-centric AD methods have been primarily studied for single-person scenarios.

It should be noted that humans naturally engage in continuous interactions with their surrounding environments and with one another. As such, anomalous human behaviors can also emerge through such multi-person interactions, where anomalies need to be jointly inspected upon multiple people. For instance, consider a scene where one person suddenly raises his/her arm. While the arm movement itself could resemble an ordinary motion (e.g., 
\textit{swinging} or \textit{exercising}), in a two-person scenario, this movement can also lead to violence, where he/she may \textit{strike} others. However, prior AD models have mostly progressed to focus on single-person behavior, and the modeling of human-human anomalies has to date remained unexplored.

\begin{figure}[t]     
    \centering     
    \includegraphics[width=1.0\linewidth]{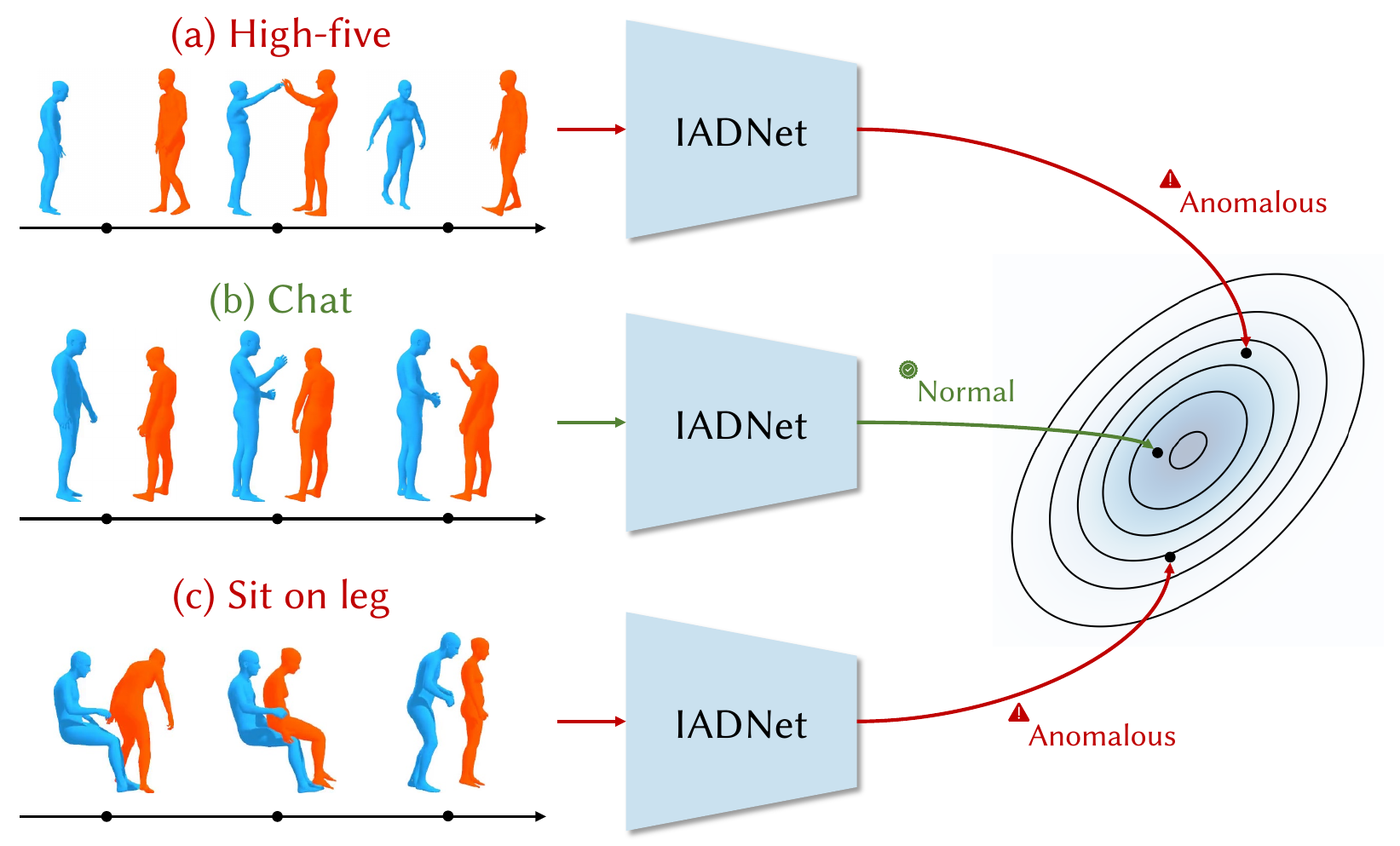}
    \caption{\textbf{Human-Human Interaction Anomaly Detection.} Given only the  \textcolor{ForestGreen}{normal} motion class selected for training (e.g., \textit{Chat}), the proposed IADNet detects whether each test interactive motion sample (a–c) is \textcolor{BrickRed}{anomalous}. }
    \label{fig:top-page}
\end{figure}  

In this paper, we comprehensively study human-centered anomalies in interactive scenes, and propose a novel benchmark: \textit{3D Human-Human Interaction Anomaly Detection (H2IAD).} As illustrated in Fig. \ref{fig:top-page}, H2IAD targets the detection of anomalies within 3D interactive motions between two individuals, learned in an unsupervised manner from only normal interactions. In particular, given a specific category of two-person interactive action sequences used as the normal class for training, the goal of H2IAD is to identify whether an arbitrary unseen interactive motion clip matches such a category. This contrasts H2IAD with traditional human action AD tasks, such as video AD \cite{flaborea2023multimodal,georgescu2021anomaly} or single-person action AD \cite{maeda2025frequency}, in that it emphasizes semantic abnormalities arising specifically from interpersonal interactions.

To address H2IAD, we propose a paradigm that aims to fully characterize the complex yet asymmetric dynamics between two interacting people. Our method is built upon a Transformer-based architecture to exploit its strong temporal encoding capacity, and is motivated by the insight that interactive motions should be modeled in a shared manner across two people to accurately represent synchronization. To this end, we introduce a Temporal Attention Sharing Module (TASM), which consists of a pair of parameter-sharing Transformer units to process each individual in parallel. Specifically, in modeling the mutual correspondence, each Transformer unit is designed to receive the query embedding from the other to capture temporal dependencies within the interaction.

Another key challenge in understanding anomalies in interactive scenes is the influence of social factors: the semantics of an interaction can drastically differ depending on spatial configurations. For example, a forward-reaching arm motion may correspond to a kind interaction, such as \textit{handing over an object}, but becomes suspicious or aggressive when the individuals are far apart. To incorporate such social cues within our model, we further introduce a Distance-Based Relational Encoding Module (DREM), which captures spatial relationships by computing dynamic pairwise distances between joints of the interacting individuals. The resulting relational embeddings are then injected into a second parameter-sharing Transformer pair within TASM, allowing the model to integrate social-spatial context alongside motion dynamics. Unlike prior multi-person motion models \cite{wen2023interactive,liu2025learning,wu2024sportshhi} that rely primarily on joint trajectories, DREM explicitly guides the network to respect interaction-level relational structures.

Given the parallel embedding streams for each interactive individual, we draw inspiration from \cite{mao2021generating} by introducing a normalizing flow (NF)-based scheme to score anomalies. The resulting Anomaly Scoring Module (ASM) learns to \textit{maximize} the likelihood of the samples within the normal category. During inference, since anomalous samples fed to the trained NF model would induce low likelihood values, we can directly employ the likelihood measurement as anomaly scores. With the above design, we dub our interaction anomaly detection model as \textbf{IADNet}, which is developed to specifically handle dynamic scenes with dense interactions. Extensive experiments on two large-scale interactive human motion datasets demonstrate that our proposed \textbf{IADNet} outperforms state-of-the-art competitors regarding detection accuracy on our newly established H2IAD task. 

We summarize our contributions as follows: 
\begin{itemize}
    \item Identifying the limitations of prior human-centered AD tasks that focus on single-person behavior, and introducing Human-Human Interaction Anomaly Detection (H2IAD), a new task that targets semantic anomalies emerging from interpersonal actions.  
    \item Proposing a novel paradigm, Interaction Anomaly Detection Network (\textbf{IADNet}), to characterize asymmetric, complex, and synchronous mutual dependencies for H2IAD.  
    \item Incorporating the modeling of spatial configurations within IADNet to tackle the inherent challenge of H2IAD by better respecting social cues.    
\end{itemize}

\begin{figure*}[t]
    \centering
    \includegraphics[width=1.0\linewidth]{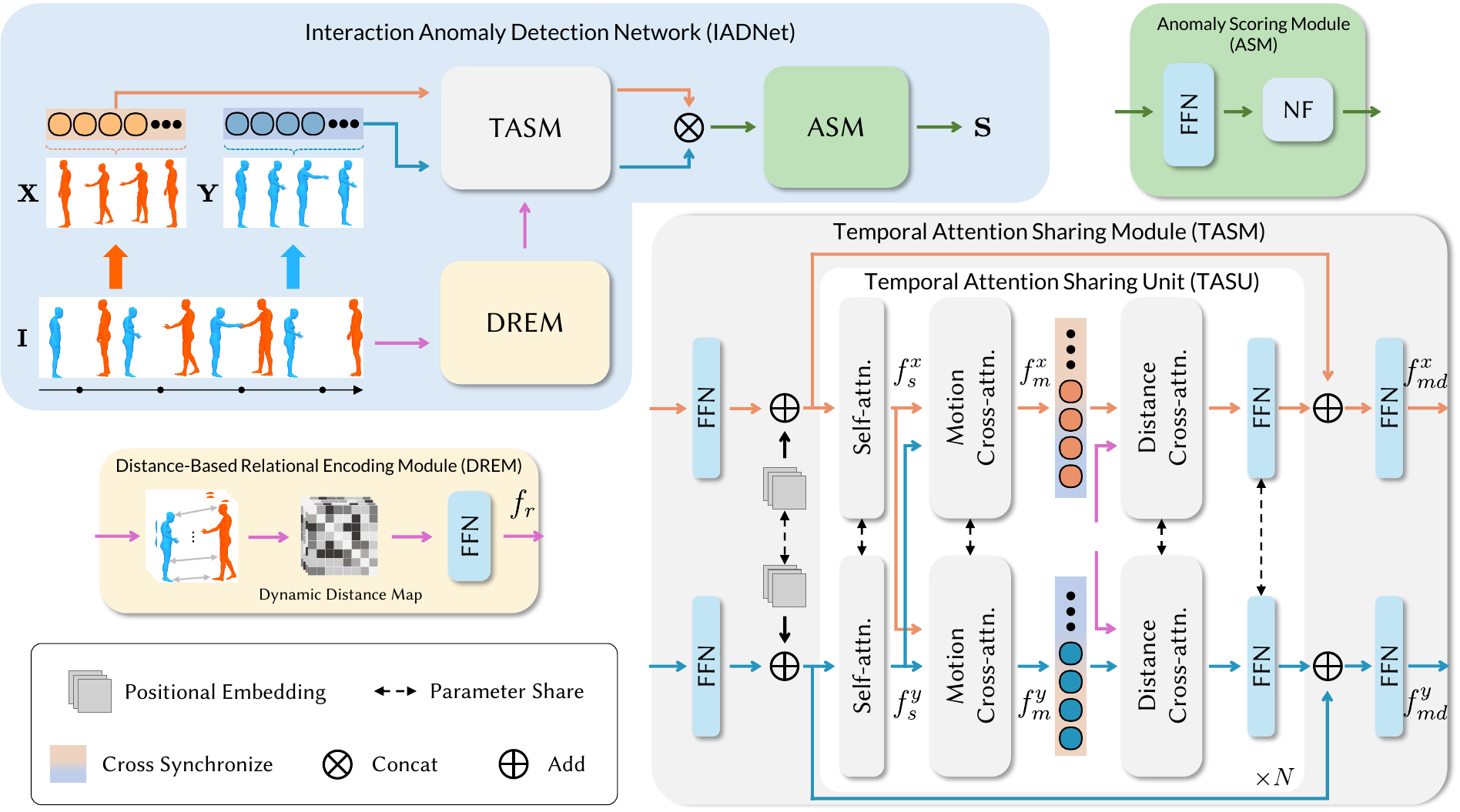}
    \caption{\textbf{Overview} of our proposed IADNet for 3D Human-Human Interaction Anomaly Detection (H2IAD). IADNet takes a pair of 3D motions as input to detect anomalous interactions with likelihood-based anomaly score measurements. }
    \label{fig:Our_approach}
\end{figure*}

\section{Related Work}

In this section, we first review previous human-related AD methods. We then discuss modeling techniques for human interaction.

\subsection{Human-Centered AD}
The task of AD has been primarily developed to date towards 2D industrial images \cite{bae2023pni,zhang2024realnet,chen2024unified,yang2025padnet,zhou2025vq}.
For human-centered AD applications, the majority of works \cite{georgescu2021anomaly,morais2019learning,georgescu2021background,wang2018video,wu2023dss,ristea2022self, zhang2024multi} aim to detect anomalies in human behaviors or events, which is typically referred to as video AD. These approaches mainly differ in their modeling frameworks for human motion, and can be primarily categorized into those applying generative paradigms, such as VAE \cite{ionescu2019object,fan2020video, karami2025graph, stergiou2024holistic} or diffusion models \cite{flaborea2023multimodal, rai2024video, tan2024frequency}, and using powerful encoding schemes, like Graph Convolutional Networks (GCNs) \cite{hirschorn2023normalizing,markovitz2020graph,flaborea2024contracting, wu2024flow} or Transformers \cite{noghre2024exploratory,ghadiya2024cross,noghre2024posewatch}. 
Generative frameworks focus on detecting anomalies via human motion reconstruction/prediction. Karami et al. \cite{karami2025graph} introduced a subgraph shuffling strategy to encourage distinguishing region-level anomalous motion. Tan et al. \cite{tan2024frequency} proposed a frequency-guided conditional diffusion framework that uses 2D-DCT to separate high- and low-frequency components of human motion to prioritize reconstruction of semantic motion. For the approaches that develop better encoding schemes, a strong temporal learning capacity can be pursued with carefully designed model architectures to pursue accurate anomaly scoring. Wu et al. \cite{wu2024flow} leveraged frame attention and skeleton attention to learn the discriminative representation of normal human motion. Flaborea et al. \cite{flaborea2024contracting} applied GCNs and learn to contract skeletal kinematic embeddings onto a compact latent hypersphere so that normal motions cluster more tightly to facilitate detection.

Overall, most human-centered AD techniques aim to identify general motion anomalies. To model anomalies within human actions at a finer level, a recent work has shifted the focus toward semantically anomalous human actions \cite{maeda2025frequency}. They introduced the first benchmark for action-level anomaly detection and proposed a multi-level action AD (ML-AAD) framework that captures both local and global motion patterns. In this sense, \cite{maeda2025frequency} shares the closest motivation to ours in detecting category-specific semantic anomalies; however, their method remains limited to single-person motion and does not account for interpersonal dynamics essential to understanding anomalous human–human interactions. Moreover, to the best of our knowledge, all existing approaches are designed exclusively for individual motion, neglecting the fact that humans inherently operate in interactive environments. Motivated by this insight, we introduce a new task, H2IAD, to learn and detect anomalous interactions between two people, and also propose a dedicated formulation to address it.

\subsection{Human Interaction Modeling}
Compared to modeling the motion of a single person, learning abstract interactions between multiple individuals is substantially more challenging. Most existing human interaction modeling techniques have been developed for motion generation, synthesis, or prediction. Due to limited availability of interactive motion data, early works approximated interpersonal interactions using a generative prior learned solely from single-person motion \cite{tevethuman}. However, such priors lack the generalization capability needed for complex multi-person scenarios. With increasing efforts dedicated to collecting large-scale interactive motion datasets, recent methods \cite{wen2023interactive, pang2022igformer, tanaka2023role, chopin2023interaction, wu2024sportshhi, faure2023holistic} directly train models on these richer data sources. Typically, these approaches employ generative frameworks equipped with specialized encoding architectures to capture inter-person dependencies. Tanaka et al. \cite{tanaka2023role} proposed a role-aware model that conditions multi-person motion synthesis on textual descriptions by explicitly modeling role assignments to encode cross-agent relations. Liang et al. \cite{liang2024intergen} introduced a cooperative denoising mechanism that encourages consistent motion embeddings between interacting individuals. Xu et al. \cite{xu2024regennet} developed an action-to-reaction synthesis pipeline using a carefully designed attention mask to enable online denoising.

Another challenge in modeling interpersonal interaction is capturing the complex social cues that arise among multiple people. This is particularly important for robotics or autonomous driving systems that demand safety-critical applications. Representative approaches mostly devise specialized graph architectures to model such dependencies. Mohamed et al. \cite{mohamed2020social} proposed a spatio-temporal graph framework to capture the social dependencies among interacting persons, such as relative velocities. Building upon it, Shi et al. \cite{shi2021sgcn} introduced a sparsity-aware graph to selectively focus on the influential neighbors to allow for context-aware motion adjustments. Zhu et al. \cite{zhu2023social} further extended these frameworks by incorporating hierarchical reasoning, allowing each agent to model not only its own motion tendencies but also to forecast the future intentions of others. Other techniques make efforts to straightforwardly model social factors, particularly spatial proximity. Pang et al. \cite{pang2022igformer} jointly learned semantic relations and spatial proximity to identify the body parts most relevant to interaction. Chopin et al. \cite{chopin2023interaction} incorporated inter-person distances and spatial context to generate interactive motions with improved realism and social plausibility.
Our method also leverages mutual distance to provide social cues during learning. However, unlike the above approaches, whose primary objective is the synthesis or prediction of interactive motion, we focus on H2IAD, which poses a conceptually different learning goal and therefore requires a more powerful design for interaction encoding.

\section{Method}
Let us now introduce our approach to H2IAD. Formally, we represent the motion sequences of the two paired interacting individuals as $\mathbf{I}^c=\{\mathbf{X}^{c},\mathbf{Y}^{c}\}$, where $\mathbf{X}^{c} = [\mathbf{x}^{c}_{1}, \cdots, \mathbf{x}^{c}_{T}]$ and $\mathbf{Y}^{c} = [\mathbf{y}^{c}_1, \cdots, \mathbf{y}^{c}_T]$ describe motions for the $c$-th interactive action category with $T$ frames. Here, $\mathbf{x}^c_i, \mathbf{y}^c_i \in \mathbb{R}^{3D}$ denote $3D$-dimensional human pose representation with $D$ joints. Given the training motion set containing $K$ samples in the $c$-th interaction category $\mathcal{I}^c=\{\mathbf{I}^c_k\}_{k\in K}$, our objective is to identify whether a testing motion clip pair $\mathbf{I}^{c_u}$ from an unseen action category $c_u$ behaves consistently with the interactive correlation within $c$. As illustrated in Fig. \ref{fig:Our_approach}, our proposed \textbf{IADNet} involves a Temporal Attention Sharing Module (TASM) for capturing inter-person motion dependencies and a Distance-Based Relational Encoding Module (DREM) to encode spatial relationships. The learned motion embedding is eventually fed to a normalizing-flow model for anomaly scoring. Below we detail our method.

\subsection{Temporal Attention Sharing Module}
The key to achieving high AD accuracy lies in how effectively the model learns the strong interactive relations. A straightforward approach is to encode each person’s actions either (a) with their own model, or (b) with one common model (e.g., Transformer) that maps both individuals’ motions into a joint feature space. We argue that either case can be less effective given the following considerations: (i) Human interactions are inherently  \textit{synchronous}. Purely independent encodings, as in (a), fail to capture such temporal coupling between participants; (ii) Human interactions also involve latent, dynamic dependencies that require \textit{collaborative} understanding of both individuals’ actions; yet, the shared encoding scheme in (b) provides no explicit mechanism to model these collaborative dynamics.

We are thus motivated to devise a dedicated encoding module to jointly address the above issues. In particular, we propose a Temporal Attention Sharing Module (TASM), which primarily consists of multiple stacked Temporal Attention Sharing Units (TASUs). Each TASU involves a pair of identically-structured encoding streams designed in parallel for each individual. Importantly, these two streams are designed to be parameter-sharing, to capture the characteristics of synchronous dynamics, and adopt the Transformer backbone to fully exploit its strong temporal modeling capacity. In particular, given the MLP-encoded action for each person, TASU first applies individual self-attention to obtain the embeddings $f^x_s$ and $f^y_s$, respectively. 

\noindent \textbf{Synchronized Positional Embedding.}
Typically, Transformer-based learning techniques involve positional encoding with fixed sinusoidal functions \cite{vaswani2017attention} to provide each token with temporal or sequential awareness. However, as H2IAD involves the modeling of interactive individuals, traditional positional encoding cannot represent the coupled temporal structure required for accurate interaction learning. To bypass this barrier, we introduce a learnable positional embedding matrix $\mathbf{H} \in \mathbb{R}^{T \times E}$ and share it across both streams for synchronization. Each row of $\mathbf{H}$ corresponds to a vector $h_t \in \mathbb{R}^E$, and $E$ denotes the embedding dimension. The motion embedding at each timestep $t$ is therefore associated with a learnable embedding vector $h_t$,  leading the input to self-attention to be the summation of the motion embedding with $\mathbf{H}$. With this shared learnable positional encoding, our self-attention modules can directly exploit rich and interaction-aware temporal representations, which contribute to improved capacity to reason jointly over the dynamics of both individuals.
We next need to discuss how to utilize such synchronized embeddings for encoding.

\noindent \textbf{Motion Cross-Attention.} To enable our model to capture mutual dependencies in the embedding space, we introduce a Motion Cross-Attention (MCA) learning stage in each TASU stream, which performs dynamic feature fusion between the interacting motions. Concretely, each MCA leverages the embedding from one person to project the query feature, and uses the embedding from the other person to derive the key and value features for cross-attention modeling. Formally, we denote query, key and value features calculated from the features of $f^{x}_{s}$ and $f^{y}_{s}$ as $\mathbf{Q}^{x}, \mathbf{Q}^{y}$, $\mathbf{K}^{x}, \mathbf{K}^{y}$, and $\mathbf{V}^{x}, \mathbf{V}^{y}$, respectively. Following the attention learning mechanism
\begin{equation}
    \mathit{Atten}(\textbf{Q},\textbf{K}, \textbf{V}) = \sigma\left( (\mathbf{Q}\cdot(\mathbf{K})^\top)/{\sqrt{C}} \right) \cdot \mathbf{V}, 
\end{equation}
MCA implements two complementary cross-attention operations: $\mathit{Atten}(\textbf{Q}^{x},\textbf{K}^{y}, \textbf{V}^{y})$ and $\mathit{Atten}(\textbf{Q}^{y},\textbf{K}^{x}, \textbf{V}^{x})$ in each stream, respectively. Here, $\sigma(\cdot)$ computes the softmax normalization, and $C$ is the dimensionality of the projection space.
\textit{In essence, our MCA drives each individual’s representation to attend to and align with its collaborative counterpart, enforcing an explicit relational awareness that reflects the dynamic and interdependent nature of human interactions. } The fused cross-synchronized feature in each stream is denoted as $f^x_{m}$ and $f^y_{m}$, respectively.

\begin{figure}
    \centering
    \includegraphics[width=0.9\linewidth]{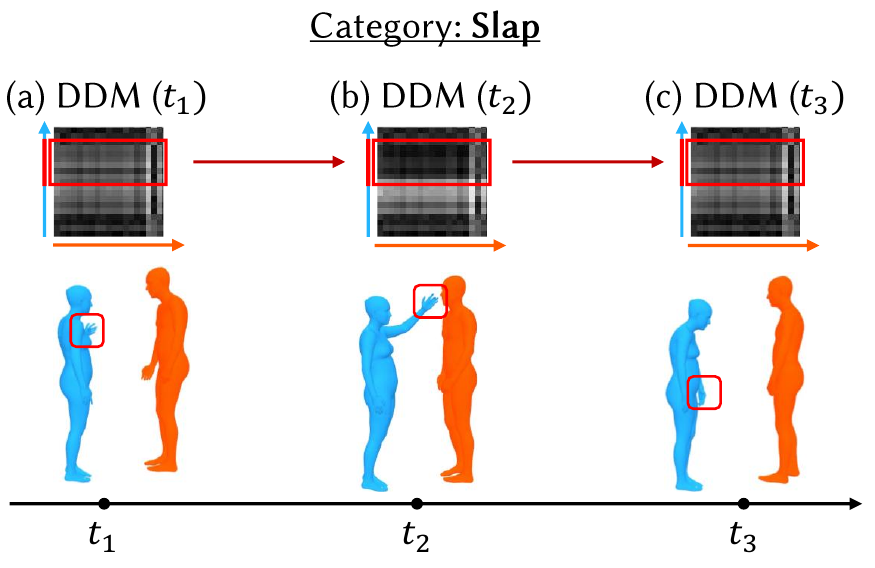}
    \caption{\textbf{An example of changes of the Dynamic Distance Map (DDM)} at different time steps $t$ for the \textit{Slap} interaction. Red boxes denote the relative distance from the joints on the \textcolor{red}{hand} of the \textcolor{cyan}{left person} to \textit{all} joints of the \textcolor{orange}{right person}. Darker pixels denote larger values (i.e., smaller distances) in DDMs. }
    \label{fig:DDM}
\end{figure}

\subsection{Distance-Based Relational Encoding Module}
While TASM effectively yields representations enriched with joint-level temporal dependencies, interaction modeling in H2IAD further requires capturing the sociality embedded in human interactions. We observe that the spatial configuration among interacting body parts across two individuals plays a decisive role in identifying interaction anomalies. For instance, in a \textit{handshake}, the relationship between the two hands is essential, whereas in a \textit{kick}, the hips and feet become the primary focus. Moreover, even within a single interaction class, the key interacting joints can shift dynamically over time. To therefore incorporate these spatial cues, we further introduce into each TASU a Distance-Based Relational Encoding Module (DREM), which explicitly learns distance-based pairwise spatial correlations between critical body parts across the two individuals.
For each training sequence pair, we prepare a distance map set $\mathcal{M}$ to include a series of dynamic distance matrices $\mathcal{M}=\{\mathbf{{M}}_t\}_{t \in T}$. Each map $\mathbf{M}_t \in \mathbb{R}^{D \times D}$ stores, in its $(i, j)$ entry, the $\mathcal{L}_2$ distance between the $i$-th joint of person $\mathbf{X}$ and the $j$-th joint of person $\mathbf{Y}$, following:
\begin{equation}
\label{Formula:IDM}
    \mathbf{m}^{(i, j)}_t = - || \mathbf{x}^{(i)}_t - \mathbf{y}^{(j)}_t ||_2.
\end{equation}
Here, we take the negative distance so that spatially closer joint pairs receive larger numerical values, allowing downstream modules to better emphasize physically interacting regions. An example of the changes of our dynamic distance map is visualized in Fig. \ref{fig:DDM}. Finally, we project the relational map sequence into the feature space with an MLP to obtain the embedding $f_{r}$, which serves as the output of our DREM. 

To incorporate these spatial cues, we introduce a distance cross-attention (DCA) learning stage in each encoding stream within our TASU, analogous to MCA. Specifically, each DCA receives queries $\mathbf{Q}^{*}_{d}$ derived from the motion-centered features $f^{*}_{m}$, while the keys and values $(\mathbf{K}^{*}_{d}, \mathbf{V}^{*}_{d})$ are obtained from $f_{r}$ ($*\in\{x,y\}$). Each DCA operation can be given by: $\mathit{Atten}(\mathbf{Q}^{*}_{d},\mathbf{K}^{*}_{d}, \mathbf{V}^{*}_{d})$, which relates motion features to their corresponding spatial-distance patterns and thus encodes dynamic joint-to-joint relational dependencies. 

Up to this point, we have introduced our \textbf{IADNet}, which primarily leverages the combination of TASM and DREM for encoding. In our implementation, the TASU is repeated for $N$ times to deepen the structure, where the learning between each TASU is realized in a residual manner. All learnable parameters in TASU are shared in both streams. By again applying an MLP, our  TASM eventually outputs the feature pair ($f^x_{md}, f^y_{md}$), which is concatenated to form $\mathbf{f}$ and passed through the anomaly scoring module (ASM).

\subsection{Normalizing Flow for Anomaly Scoring.}
Inspired by \cite{maeda2025frequency,hirschorn2023normalizing}, our ASM applies the Normalizing Flow (NF) model to accurately characterize the distribution of interaction features. A key advantage of NF is its ability to compute the exact likelihood of each input, in contrast to models such as VAEs that optimize a variational upper-bound, which aligns well with the aim of AD. Formally, the NF learns a bijective transformation $\phi$ that maps the combined motion embedding $\mathbf{f}$ into a latent variable $\mathbf{s}$ following a simple Gaussian prior: $\mathbf{s} = \phi(\mathbf{f})$, where $\mathbf{s} \sim {\mathcal{N}}(0, \textbf{I})$. Here, the mapping $\phi$ is composed of a series of deep neural networks to pursue strong expressiveness. The training objective of NF is to \textit{maximize} the likelihood of each sample, which is expressed as
\begin{equation}
    p(\mathbf{f}) = g(\mathbf{s}) \left| \det \frac{\partial \phi}{\partial \mathbf{f}} \right|. 
\end{equation}
Here, $g(\mathbf{s}) = \mathcal{N}(\mathbf{s} | 0, \mathbf{I})$, and $\left| \text{det}\frac{\partial \phi}{\partial{\mathbf{f}}} \right|$ measures the absolute value of the determinant of the Jacobian of $\phi$. Since H2IAD follows the one-class classification setting, we train the NF using only the features $\mathbf{f}^c$ from the selected normal interaction class $c$ by \textit{minimizing} the negative log-likelihood (NLL):
\begin{align}
    \label{Formula: loss log-likelihood}
    \mathcal{L} & = -\log p(\mathbf{f}^c) \\
    & =-\log g(\mathbf{s}^c)-\log \left|\operatorname{det}\frac{\partial \phi}{\partial \mathbf{f}^c}\right|,
\end{align} 
where $\mathbf{s}^c = \phi(\mathbf{f}^c)$ and $\mathbf{s}^c \sim \mathcal{N}(0,\mathbf{I})$.

In our NF implementations, we follow \cite{mao2021generating,maeda2025frequency} by employing a lightweight 10-layer MLP architecture for training stability. To ensure the invertibility of $\phi$, we adopt the QR decomposition to compute the MLP weights and use the PReLU activation to preserve monotonicity. During inference, for any arbitrary test interaction features that deviate from the learned normal-class distribution, the trained NF model will produce a lower likelihood (i.e., a higher NLL) to enable a reliable detection of anomalous interaction behaviors. We thus directly employ the NLL as the anomaly score for detection.

\section{Experiment}
In this section, we report a series of experimental results to evaluate the effectiveness of our \textbf{IADNet} on our newly introduced task H2IAD against other baselines.

\begin{table*}[]
\caption{\textbf{Quantitative evaluation on Inter-X and NTU RGB+D 120}. The best AUC scores are highlighted in bold.}
    \centering    \begin{tabular}{cccccccccccccc}
    \toprule
    \multirow{2}{*}{Method}& \multicolumn{10}{c}{Inter-X}&\multirow{2}{*}{Avg.}\\
    \cmidrule(lr){2-11} 
         & \textit{Hug}& \textit{Hndsk}& \textit{Wave} & \textit{Grab} & \textit{Hit} & \textit{Kick}& \textit{Posing} & \textit{Push} & \textit{Pull}& \textit{Sit o leg} &\\
         \hline
         STG-NF \cite{hirschorn2023normalizing}  & 0.647& 0.563& 0.564& 0.612& 0.492& 0.456& 0.549& 0.688& 0.548& 0.411& \\
         MoCoDAD \cite{flaborea2023multimodal}   & 0.482& 0.555& 0.445& 0.528& 0.498& 0.450& 0.493& 0.596& 0.548& 0.461& \\
          ML-AAD \cite{maeda2025frequency} & 0.692& 0.733& 0.715& 0.645& 0.649& 0.761& \textbf{0.611}& 0.735& 0.620& 0.674& \\
         IADNet & \textbf{0.796}& \textbf{0.777}& \textbf{0.752}& \textbf{0.782}& \textbf{0.773}& \textbf{0.857}& 0.538& \textbf{0.822}& \textbf{0.707}& \textbf{0.765}& \\
         \hline
         & \textit{Slap}& \textit{Pat o bk}& \textit{Pnt fng at} & \textit{Wlk twds} & \textit{Knk ov} & \textit{Stp o ft}& \textit{High-five} & \textit{Chase} & \textit{Wsp i er}& \textit{Spt w hnd} &\\
         \hline
         STG-NF \cite{hirschorn2023normalizing}  & 0.637& 0.670& 0.401& 0.398& 0.473& 0.460& 0.601& 0.307& 0.523& 0.539& \\
         MoCoDAD \cite{flaborea2023multimodal}   & 0.609& 0.484& 0.531& 0.518& 0.458& 0.483& 0.474& 0.518& 0.428& 0.459& \\
          ML-AAD \cite{maeda2025frequency} & 0.690& 0.736& 0.599& \textbf{0.661}& 0.701& 0.692& 0.685& 0.514& 0.524& 0.577& \\
         IADNet & \textbf{0.777}& \textbf{0.756}& \textbf{0.671}& 0.652& \textbf{0.787}& \textbf{0.743}& \textbf{0.846}& \textbf{0.604}& \textbf{0.568}& \textbf{0.595}& \\
         \hline
         & \textit{Rk-pr-sr}& \textit{Dance}& \textit{Link arms} & \textit{Sld t sld} & \textit{Bend} & \textit{Cry o bk}& \textit{Msg sld} & \textit{Msg leg} & \textit{Hnd wstl}& \textit{Chat} &\\
         \hline
         STG-NF \cite{hirschorn2023normalizing}  & 0.249& 0.636& 0.616& 0.619& 0.518& 0.477& 0.529& 0.228& 0.351& 0.397& \\
         MoCoDAD \cite{flaborea2023multimodal}   & 0.545& 0.609& 0.485& 0.464& 0.465& 0.501& 0.507& 0.496& 0.494& 0.544& \\
          ML-AAD \cite{maeda2025frequency} & 0.444& 0.543& 0.726& 0.668& 0.677& 0.539& 0.575& 0.507& 0.650& 0.574& \\
         IADNet & \textbf{0.566}& \textbf{0.718}& \textbf{0.788}& \textbf{0.709}& \textbf{0.763}& \textbf{0.654}& \textbf{0.628}& \textbf{0.646}& \textbf{0.718}& \textbf{0.662}& \\
         \hline
         & \textit{Pat o chk}& \textit{Thb up}& \textit{Tch hd} & \textit{Imitate} & \textit{Ks o ck} & \textit{Help up}& \textit{Cvr mth} & \textit{Look back} & \textit{Block}& \textit{Fly kiss} &\\
         \hline
         STG-NF \cite{hirschorn2023normalizing}  & 0.667& 0.316& 0.646& 0.471& 0.626& 0.410& 0.641& 0.524& 0.522& 0.484& 0.512\\
         MoCoDAD \cite{flaborea2023multimodal}   & 0.524& 0.580& 0.549& 0.521& 0.473& 0.467& 0.564& 0.550& 0.436& 0.529& 0.508\\
          ML-AAD \cite{maeda2025frequency} & 0.709& 0.567& 0.699& 0.440& 0.709& 0.449& 0.610& 0.653& 0.522& 0.578& 0.626\\
         IADNet & \textbf{0.738}& \textbf{0.663}& \textbf{0.752}& \textbf{0.585}& \textbf{0.749}& \textbf{0.584}& \textbf{0.721}& \textbf{0.701}& \textbf{0.657}& \textbf{0.695}& \textbf{0.707}\\
         \bottomrule
    \end{tabular}
    \begin{tabular}{cccccccccccc}
    \multirow{2}{*}{Method}& \multicolumn{9}{c}{NTU RGB+D 120}&\multirow{2}{*}{Avg.}\\
    \cmidrule(lr){2-10} 
         & \textit{Punch}& \textit{Kick}& \textit{Push}& \textit{Pat o bk}& \textit{Pnt fng at}& \textit{Hug}& \textit{Give smtg}& \textit{Touch pkt}& \textit{Hndsk}&\\
         \hline
         STG-NF \cite{hirschorn2023normalizing}  & 0.458& 0.498& 0.501& 0.496& 0.497& 0.572& 0.548& 0.506& 0.531& \\
         MoCoDAD \cite{flaborea2023multimodal}   & 0.512& 0.421& 0.498& 0.499& 0.488& 0.508& 0.502& 0.511& 0.448&\\
          ML-AAD \cite{maeda2025frequency}          & 0.465& 0.495& 0.516& 0.419& 0.465& 0.571& 0.460& 0.411& 0.501&\\
         IADNet                                     & \textbf{0.593}& \textbf{0.611}& \textbf{0.661}& \textbf{0.808}& \textbf{0.882}& \textbf{0.657}& \textbf{0.760}& \textbf{0.795}& \textbf{0.882}&\\
         \hline
         & \textit{Wlk twds}& \textit{Wlk apt}& \textit{Hit w smtg}& \textit{Wld knf twds}& \textit{Knk ov}& \textit{Grab}& \textit{Shot w Gun}& \textit{Stp o ft}& \textit{Hight-five}&\\
         \hline
         STG-NF \cite{hirschorn2023normalizing}  & 0.571& 0.597& 0.492& 0.512& 0.495& 0.469& 0.484& 0.492& 0.515&\\
         MoCoDAD \cite{flaborea2023multimodal}   & 0.480& 0.489& 0.506& 0.519& 0.503& 0.493& 0.513& 0.543& 0.518&\\
          ML-AAD \cite{maeda2025frequency}          & 0.586& 0.499& \textbf{0.609}& \textbf{0.570}& 0.580& 0.463& 0.578& 0.561& \textbf{0.594}&\\
         IADNet                                     & \textbf{0.728}& \textbf{0.742}& 0.434& 0.551& \textbf{0.688}& \textbf{0.569}& \textbf{0.606}& \textbf{0.603}& 0.548&\\
         \hline
         & \textit{Chrs a drk}& \textit{Cary smtg}& \textit{Take a pht}& \textit{Follow}& \textit{Wsp i er}& \textit{Exchg thgs}& \textit{Spt w hnd}& \textit{Rk-pr-sr}& &\\
         \hline
         STG-NF \cite{hirschorn2023normalizing}  & 0.460& 0.543& 0.510& 0.498& 0.531& 0.486& 0.555& 0.482& & 0.511\\
         MoCoDAD \cite{flaborea2023multimodal}   & 0.529& 0.508& 0.489& 0.528& 0.506& 0.493& 0.491& 0.500& & 0.500\\
          ML-AAD \cite{maeda2025frequency}          & 0.543& 0.537& 0.572& 0.488& \textbf{0.577}& 0.516& 0.495& 0.486& & 0.521\\
         IADNet                                     & \textbf{0.728}& \textbf{0.599}& \textbf{0.772}& \textbf{0.614}& 0.564& \textbf{0.658}& \textbf{0.850}& \textbf{0.664}& & \textbf{0.676}\\
         \bottomrule
    \end{tabular}
    \label{Table.quantitative_AUCs}
\end{table*}

\noindent \textbf{Dataset.} Following previous works on human interaction \cite{xu2024regennet,wang2025timotion}, we evaluate our method on two large-scale interaction datasets: Inter-X \cite{xu2024inter} and NTU RGB+D 120 \cite{liu2019ntu}.

\begin{itemize}
    \item \textbf{Inter-X} \cite{xu2024inter} is so far the largest public human-human interaction dataset featuring an extensive range of interaction patterns. It comprises 11,388 interaction sequences and records over 8.1 million frames, capturing diverse scenarios where multiple individuals perform various actions, such as \textit{Handshaking}, \textit{High-fiving}, and \textit{Fighting}. We use the officially provided training and testing split by the dataset (9,110 for training and 2,278 for testing). The dataset covers 40 distinct interaction categories, and the human pose in each frame is represented by 64 joints with 3D coordinates. 
    \item  \textbf{NTU RGB+D 120} \cite{liu2019ntu} is a large-scale human motion dataset that contains 26 categories of interactive actions. It provides 8,118 human–human interaction sequences captured from three camera viewpoints, covering a broad range of daily activities such as \textit{Hugging}, \textit{Walking Toward}, and \textit{Following}. Following \cite{xu2024regennet}, we use only the data captured by camera 1 and adopt the standard cross-subject evaluation protocol, where half of the subjects are used for training and the remaining subjects for testing. Each human pose is represented as a 56-joint 3D skeleton.
\end{itemize}

\noindent\textbf{Baselines.} Since there is no prior work tackling the interaction-specific AD task we propose, we adapt the state-of-the-art video AD models, STG-NF \cite{hirschorn2023normalizing}, MoCoDAD \cite{flaborea2023multimodal}, and the action AD method, ML-AAD  \cite{maeda2025frequency} to our task.
Specifically, STG-NF uses a spatio-temporal Graph Convolutional Network (ST-GCN) to encode 2D pose sequences, which we adapt to take 3D location during learning. Similarly, the diffusion model of MoCoDAD is also retrained to learn from 3D representations.

For each training sample involving two interacting individuals, we prepare and train two separate models, with each learning the motion patterns of one person. Specifically, for every interaction category, these models independently learn each individual’s actions as the normal class, resulting in two independently trained models per category. During inference, the action of each person is input into its corresponding model to compute an AUC value. We choose the higher value from the two obtained AUCs as the final result.

\noindent \textbf{Evaluation Metrics.} Following the one-class classification protocol as prior AD methods \cite{hirschorn2023normalizing, markovitz2020graph, flaborea2023multimodal}, we employ the Area Under the Receiver Operating Characteristic Curve (AUROC) to evaluate our method. The ROC curve depicts the trade-off between the True Positive Rate (TPR, $\text{TPR} = TP / (TP + FN)$) and the False Positive Rate (FPR, $\text{FPR} = FP / (FP + TN)$) across different decision thresholds. Here, $\text{TP}$, $\text{FN}$, $\text{FP}$, and $\text{TN}$ represent the number of true positive instances, false negatives, false positives, and true negatives, respectively. A higher AUROC value indicates a stronger capability to effectively distinguish normal interactions from anomalous ones.

\begin{figure*}
    \centering
    \begin{minipage}{0.48\textwidth}
        \centering
        \includegraphics[width=1.0\linewidth]{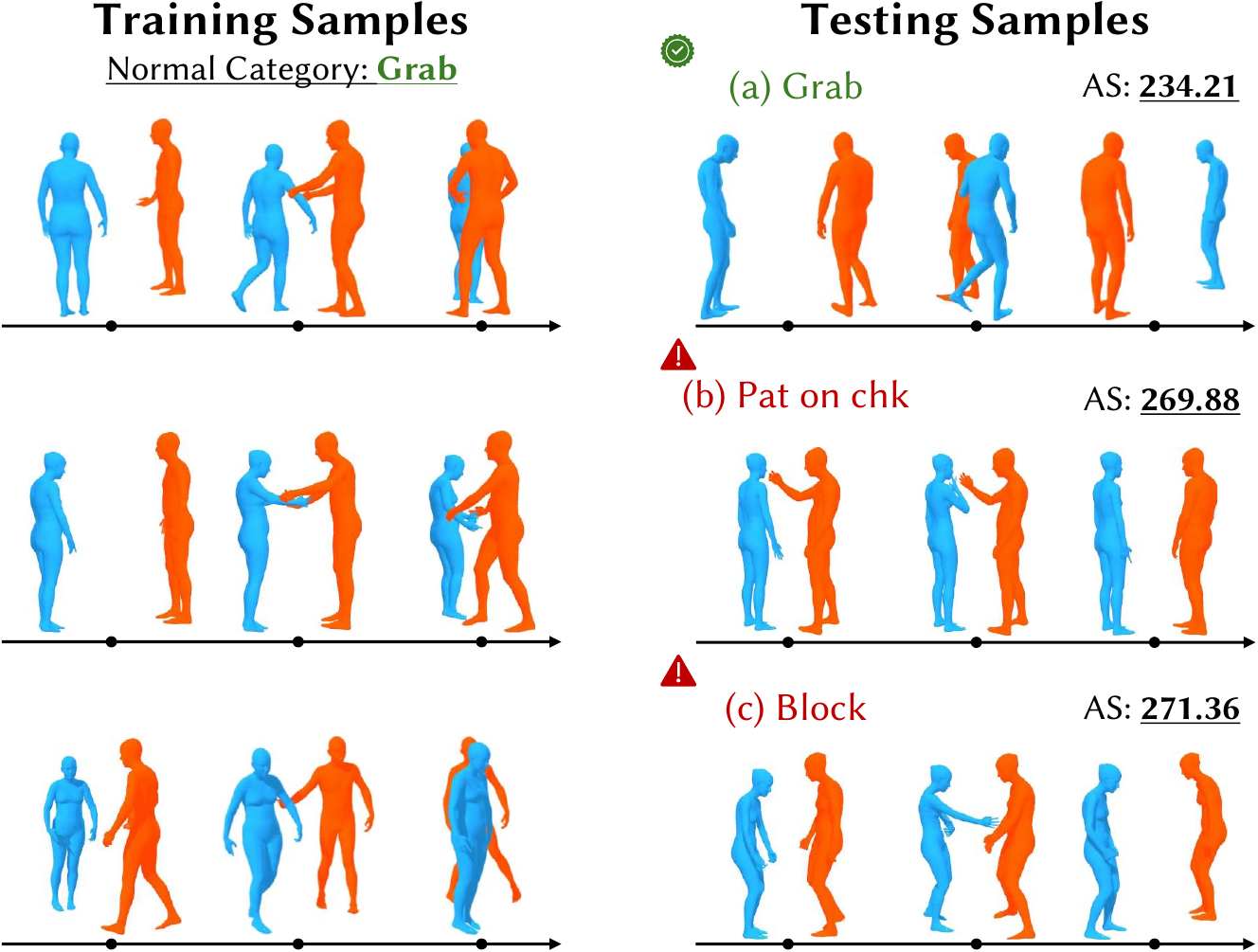}
    \end{minipage}
    \hfill
    \begin{minipage}{0.48\textwidth}
        \centering
        \includegraphics[width=1.0\linewidth]{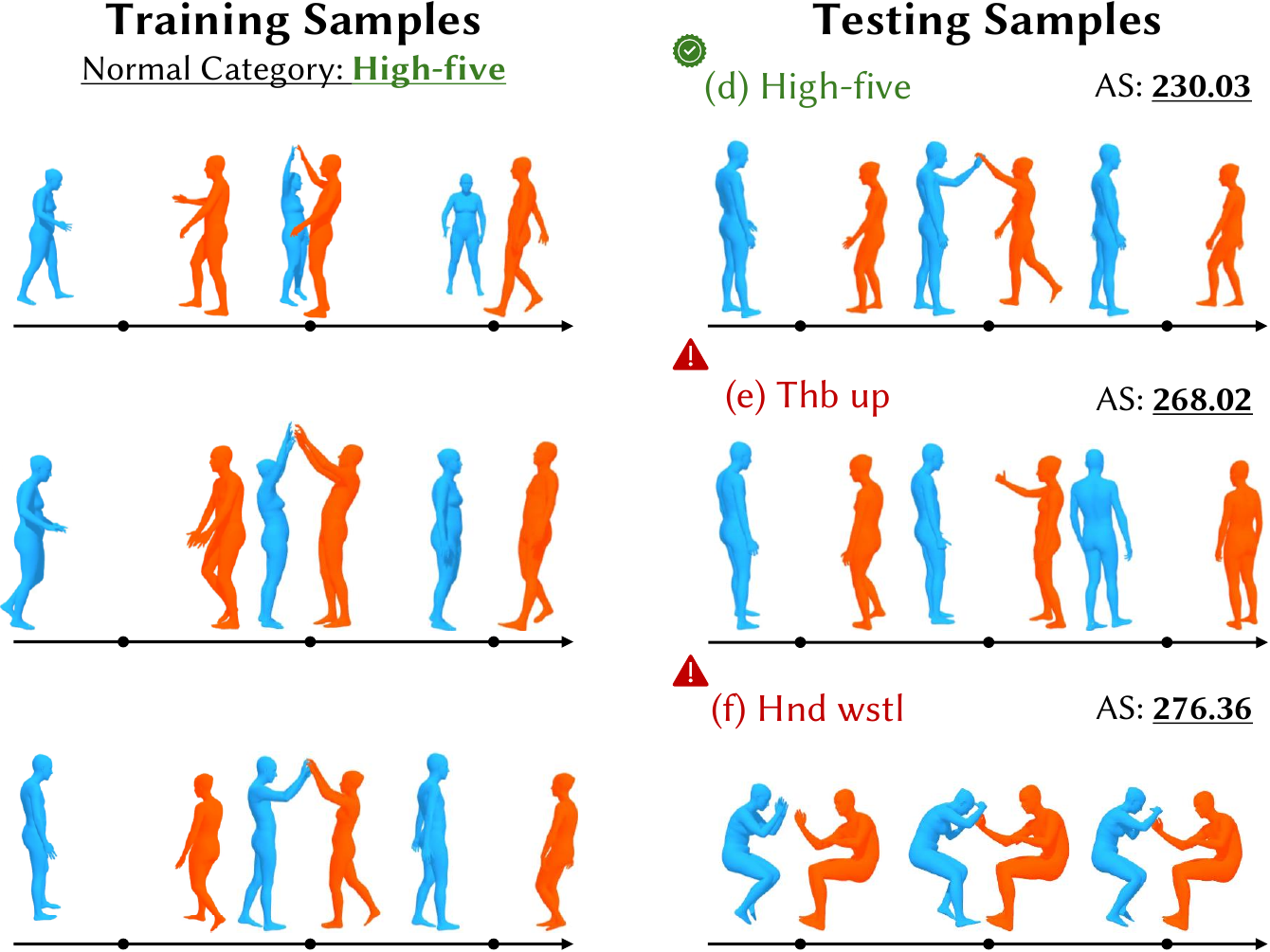}
    \end{minipage}
    \caption{\textbf{Qualitative results} with two different normal interaction category settings (\textit{Grab} and \textit{High-five}) on Inter-X.  For each case, we show the training samples on the left side, and the testing samples (a-c), (d-f) on the right. ``AS'' refers to the anomaly score (i.e., negative log-likelihood) for each sample during inference. A higher AS denotes a stronger abnormality. }
    \label{fig:result_inter-X}
\end{figure*}

\begin{figure*}
    \centering
    \begin{minipage}{0.48\textwidth}
        \centering
        \includegraphics[width=1.0\linewidth]{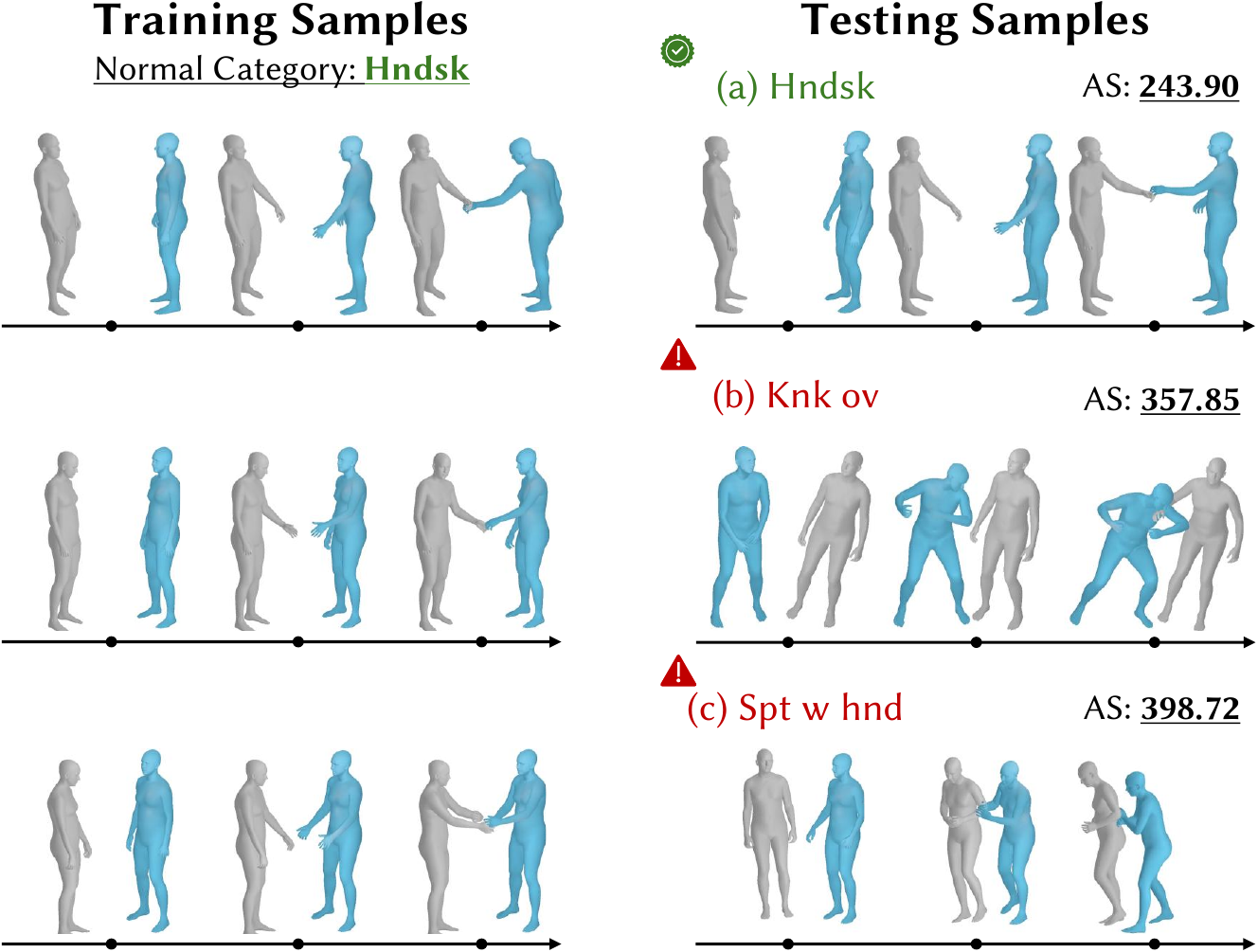}
    \end{minipage}
    \hfill
    \begin{minipage}{0.48\textwidth}
        \centering
        \includegraphics[width=1.0\linewidth]{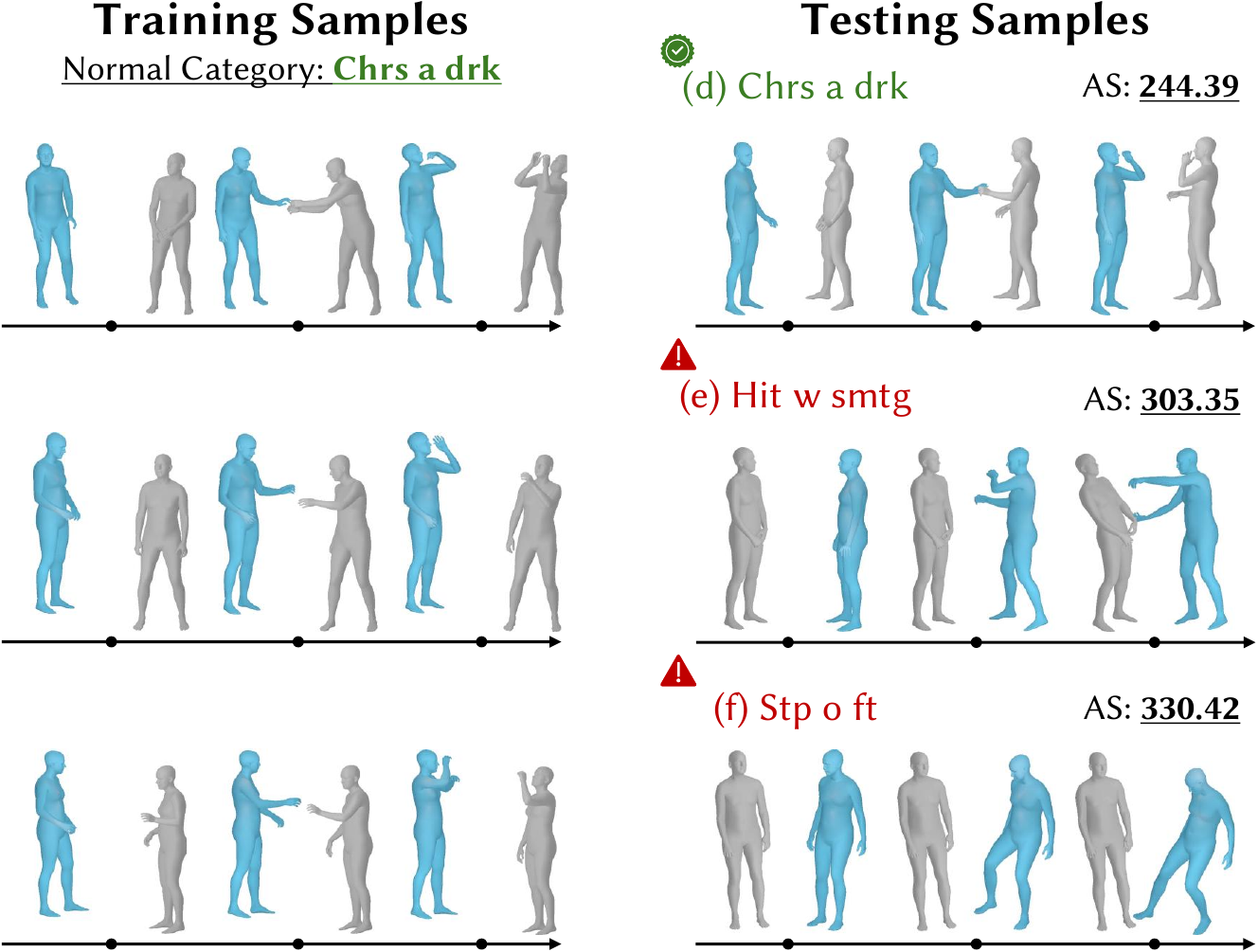}
    \end{minipage}
    \caption{\textbf{Qualitative results} with two different normal interaction category settings (\textit{Handshake} and \textit{Cheers and drink}) on NTU RGB+D 120.  For each case, we show the training samples on the left side, and the testing samples (a-c), (d-f) on the right. ``AS'' refers to the anomaly score (i.e., likelihood) for each sample during inference. A higher AS denotes a stronger abnormality. }
    \label{fig:result_NTU}
\end{figure*}

\noindent \textbf{Implementation Details.} We implement our model in Pytorch and train it with the Adam \cite{kingma2014adam} optimizer on RTX3090. We set the initial learning rate to 0.001 and decay it to 1e-5 by the 50th epoch to ensure stable convergence. We train our model for 50 epochs for each interaction category on both datasets. The number of TASU $N$ is set to 8, with a hidden size of 512. 

\subsection{Results}

\noindent \textbf{Quantitative Results.} We compare in Tabs. \ref{Table.quantitative_AUCs} our results against those of baselines on Inter-X and NTU RGB+D 120, respectively. For each column in both tables, all models, including ours and the baselines, are retrained using only the corresponding interactive action as the normal class. It can be confirmed that our proposed method outperforms the baselines on almost all action categories on both datasets. Notably, IADNet outperforms STG-NF \cite{hirschorn2023normalizing} and MoCoDAD \cite{flaborea2023multimodal} by a large margin on both datasets, which suggests that these two methods are less effective at capturing and understanding human behaviors with interactions.

Let us now focus on ML-AAD \cite{maeda2025frequency}, which, although designed for single-person AD, represents the state of the art most closely aligned with our goal of modeling semantic human actions. While its multi-level strategy effectively models category-specific motion characteristics, leading to better results than the other baselines, our method still achieves higher AUC scores in most categories across both datasets. Notably, ML-AAD struggles on actions that involve strong interpersonal interaction. For example, for categories such as \textit{Dance} and \textit{Help up} in Inter-X, the accuracy of ML-AAD drops significantly. This is primarily because, despite ML-AAD’s ability to model semantic anomalies, it lacks a learning mechanism for inter-personal dependencies and therefore cannot fully capture the underlying interaction dynamics. In contrast, our IADNet, with its TASM module, successfully models these implicit relationships, yielding accuracy improvements of 17.5\% on \textit{Dance} and 13.5\% on \textit{Help up}. Similar observations can be found in NTU RGB+D 120, for interaction categories like \textit{Give something} and \textit{Rock-paper-scissors}. These results demonstrate the effectiveness of IADNet for H2IAD.

We next study a failure case of our method. As shown in Tab. \ref{Table.quantitative_AUCs} on NTU RGB+D 120, the detection accuracy via our IADNet on the \textit{Hit with something} still leaves room for improvement. Precisely, the interaction within \textit{Hit with something} involves striking another person using a handheld external object. However, since NTU RGB+D 120 does not include the object models, such human-object interactive seniors cannot be fully captured. Consequently, the resulting motion pattern can appear highly similar to object-free actions, such as \textit{Punch}, which leads to a more challenging discrimination and thereby degrades accuracy. Addressing this issue can entail the development of new datasets that jointly involve human-human and human-object interactions, which we would like to explore in the future.


\begin{table*}
\caption{\textbf{Effect of different positional encoding schemes} on Inter-X with AUC. The best AUC scores are denoted in bold.}
    \centering    
    \begin{tabular}{cccccccccccc}
         \toprule
         \multirow{2}{*}{Method}& \multicolumn{10}{c}{Inter-X}&\multirow{2}{*}{Avg.}\\
         \cmidrule(lr){2-11} 
         & \textit{Hug}& \textit{Hndsk}& \textit{Wave} & \textit{Grab} & \textit{Hit} & \textit{Kick}& \textit{Posing} & \textit{Push} & \textit{Pull}& \textit{Sit o leg} &\\
         \hline
         Sinusoidal          & 0.700& 0.629& 0.629& 0.738& 0.710& 0.772& \textbf{0.538}& 0.815& 0.705& 0.583& \\
         Unsync. embedding & 0.747& 0.721& 0.736& 0.767& 0.722& 0.758& 0.534& \textbf{0.849}& 0.679& 0.610& \\
         Sync. embedding & \textbf{0.796}& \textbf{0.777}& \textbf{0.752}& \textbf{0.782}& \textbf{0.773}& \textbf{0.857}& \textbf{0.538}& 0.822& \textbf{ 0.707}& \textbf{0.765}& \\
         \hline
         & \textit{Slap}& \textit{Pat o bk}& \textit{Pnt fng at} & \textit{Wlk twds} & \textit{Knk ov} & \textit{Stp o ft}& \textit{High-five} & \textit{Chase} & \textit{Wsp i er}& \textit{Spt w hnd} &\\
         \hline
         Sinusoidal          & 0.771& 0.742& 0.646& \textbf{0.657}& 0.715& 0.651& 0.536& \textbf{0.626}& 0.546& 0.573& \\
         Unsync. embedding & 0.760& 0.708& 0.619& 0.637& 0.771& 0.588& 0.787& 0.580& 0.468& \textbf{0.639}& \\
         Sync. embedding & \textbf{0.777}& \textbf{0.756}& \textbf{0.671}& 0.652& \textbf{0.787}& \textbf{0.743}& \textbf{0.846}& 0.604& \textbf{0.568}& 0.595& \\
         \hline
         & \textit{Rk-pr-sr}& \textit{Dance}& \textit{Link arms} & \textit{Sld t sld} & \textit{Bend} & \textit{Cry o bk}& \textit{Msg sld} & \textit{Msg leg} & \textit{Hnd wstl}& \textit{Chat} &\\
         \hline
         Sinusoidal          & 0.471& 0.558& 0.727& 0.604& \textbf{0.781}& 0.493& 0.483& 0.493& 0.676& 0.505& \\
         Unsync. embedding & 0.552& 0.570& 0.665& 0.656& 0.767& 0.620& 0.492& 0.537& 0.489& 0.659& \\
         Sync. embedding & \textbf{0.566}& \textbf{0.718}& \textbf{0.788}& \textbf{0.709}& 0.763& \textbf{0.654}& \textbf{0.628}& \textbf{0.646}& \textbf{0.718}& \textbf{0.662}& \\
         \hline
         & \textit{Pat o chk}& \textit{Thb up}& \textit{Tch hd} & \textit{Imitate} & \textit{Ks o ck} & \textit{Help up}& \textit{Cvr mth} & \textit{Look back} & \textit{Block}& \textit{Fly kiss} &\\
         \hline
         Sinusoidal          & 0.687& 0.542& \textbf{0.755}& 0.378& 0.726& 0.569& 0.681& \textbf{0.716}& 0.554& 0.657& 0.633\\
         Unsync. embedding & 0.721& 0.565& 0.732& 0.501& 0.743& 0.578& 0.697& 0.700& 0.647& \textbf{0.702}& 0.657\\
         Sync. embedding & \textbf{0.738}& \textbf{0.663}& 0.752& \textbf{0.585}& \textbf{0.749}& \textbf{0.584}& \textbf{0.721}& 0.701& \textbf{0.657}& 0.695& \textbf{0.707}\\
         \bottomrule
    \end{tabular}   
    \label{Table. Inter-X Positional Encoder AUC}
\end{table*}

\begin{table*}
\caption{\textbf{Effect of DREM} during training on Inter-X with AUC. Higher AUC scores are denoted in bold.}
    \centering    \begin{tabular}{cccccccccccccc}
    \toprule
    \multirow{2}{*}{Method}& \multicolumn{10}{c}{Inter-X}&\multirow{2}{*}{Avg.}\\
    \cmidrule(lr){2-11} 
         & \textit{Hug}& \textit{Hndsk}& \textit{Wave} & \textit{Grab} & \textit{Hit} & \textit{Kick}& \textit{Posing} & \textit{Push} & \textit{Pull}& \textit{Sit o leg} &\\
         \hline
         w/o DREM & 0.713& 0.748& 0.651& 0.679& 0.746& 0.823& 0.527& 0.799& 0.615& 0.680& \\
         w. DREM & \textbf{0.796}& \textbf{0.777}& \textbf{0.752}& \textbf{0.782}& \textbf{0.773}& \textbf{0.857}& \textbf{0.538}& \textbf{0.822}& \textbf{0.707}& \textbf{0.765}& \\
         \hline
         & \textit{Slap}& \textit{Pat o bk}& \textit{Pnt fng at} & \textit{Wlk twds} & \textit{Knk ov} & \textit{Stp o ft}& \textit{High-five} & \textit{Chase} & \textit{Wsp i er}& \textit{Spt w hnd} &\\
         \hline
         w/o DREM & 0.725& 0.697& 0.621& 0.648& 0.751& 0.720& 0.768& 0.576& 0.566& 0.512& \\
         w. DREM & \textbf{0.777}& \textbf{0.756}& \textbf{0.671}& \textbf{0.652}& \textbf{0.787}& \textbf{0.743}& \textbf{0.846}& \textbf{0.604}& \textbf{0.568}& \textbf{0.595}& \\
         \hline
         & \textit{Rk-pr-sr}& \textit{Dance}& \textit{Link arms} & \textit{Sld t sld} & \textit{Bend} & \textit{Cry o bk}& \textit{Msg sld} & \textit{Msg leg} & \textit{Hnd wstl}& \textit{Chat} &\\
         \hline
         w/o DREM & 0.431& 0.555& 0.701& 0.643& \textbf{0.764}& 0.546& 0.507& 0.370& \textbf{0.731}& 0.459& \\
         w. DREM & \textbf{0.566}& \textbf{0.718}& \textbf{0.788}& \textbf{0.709}& 0.763& \textbf{0.654}& \textbf{0.628}& \textbf{0.646}& 0.718& \textbf{0.662}& \\
         \hline
         & \textit{Pat o chk}& \textit{Thb up}& \textit{Tch hd} & \textit{Imitate} & \textit{Ks o ck} & \textit{Help up}& \textit{Cvr mth} & \textit{Look back} & \textit{Block}& \textit{Fly kiss} &\\
         \hline
         w/o DREM & 0.711& 0.531& 0.706& 0.413& \textbf{0.758}& 0.558& 0.719& 0.686& 0.602& 0.687& 0.641\\
         w. DREM & \textbf{0.738}& \textbf{0.663}& \textbf{0.752}& \textbf{0.585}& 0.749& \textbf{0.584}& \textbf{0.721}& \textbf{0.701}& \textbf{0.657}& \textbf{0.695}& \textbf{0.707}\\
         \bottomrule
    \end{tabular}

    \label{Table. Inter-X Distances AUCs}
\end{table*}

\begin{table}
\caption{\textbf{Analysis on the distance} for representative interaction categories on Inter-X. For each interaction category, we report: (i) the average inter-joint displacement $dsp$, (ii) the AUC obtained with DREM, and (iii) the AUC improvement $\Delta$AUC over the model without DREM.}
    \centering    \begin{tabular}{cccc}
    \toprule
    Category& \textit{dsp}& AUC (w. DREM)& $\Delta$AUC \\ \hline
    \textit{Imitate}&  7.13& 0.585& +0.172\\ 
    \textit{Massaging leg}&  5.74& 0.646& +0.276\\ 
    \textit{Dance}&  5.65& 0.718& +0.163\\
    \hdashline
    \textit{Fly kiss}& 4.50& 0.695& +0.008\\ 
    \bottomrule
    \end{tabular}
    \label{Table. Inter-X Distances Value}
\end{table}

\noindent{\textbf{Qualitative Results.}} 
To further illustrate the effectiveness of our method on H2IAD, we provide qualitative detection results for Inter-X and NTU RGB+D 120 in Fig. \ref{fig:result_inter-X} and Fig. \ref{fig:result_NTU}, respectively. For each example, the left side visualizes several normal interaction samples used during training, while the right side presents three test samples from different categories along with their corresponding anomaly scores (AS). We can observe that our IADNet consistently achieves a noticeable NLL increase (i.e.,  likelihood drop) for anomalous samples. In particular, although the \textit{Cheers and drinks} interaction in Fig. \ref{fig:result_NTU}(d) primarily involves coordinated hand motions, samples exhibiting active lower-body movements (e.g., \textit{Stop on foot} in Fig. \ref{fig:result_inter-X}(d)) trigger a significant NLL rise of 86.03. Moreover, even for visually similar interactions involving hand gestures, such as \textit{Highfive} and \textit{Thumb up}  in Fig. \ref{fig:result_NTU}(d) and (e), IADNet still manages to distinguish their semantic differences. We assume that this is due to our DREM, which incorporates spatial cues to socially distinguish subtle anomalies for similar movements. Overall, above analysis validates that IADNet effectively detects diverse types of anomalous human interactions, including (i) local anomalies within specific body parts and (ii) visually similar actions that differ semantically.

\subsection{Ablation Study}
To gain deeper insights into our model, we evaluate the influence of the following main components:

\noindent{\textbf{Effect of Positional Encoding.}}
We apply synchronized positional embeddings across both streams in TASU to incorporate sequential cues into the Transformer. To evaluate the effectiveness, we compare our design with an unsynchronized variant and with the standard sinusoidal positional encoding, and summarize the results on Inter-X in Tab. \ref{Table. Inter-X Positional Encoder AUC}. We can see that our synchronized scheme consistently achieves superior performance across most categories. In particular, notable improvements are observed in interaction types such as \textit{High-five} and \textit{Sit on leg}, where precise temporal alignment and coordinated responses are crucial. Overall, these results demonstrate the capability of our synchronized positional embedding to better capture both symmetric and asymmetric interpersonal dynamics.

\noindent{\textbf{Effect of DREM.}}
Our IADNet incorporates the DREM to introduce distance-based correspondence to the model to capture social cues within interpersonal dynamics. To understand the effectiveness of DREM in our framework, we perform an ablation study on DREM and show the results in Tab. \ref{Table. Inter-X Distances AUCs}. We can see in Tab. \ref{Table. Inter-X Distances AUCs} that integrating DREM generally leads to higher AUC scores across most interaction categories. In particular, remarkable gains are observed in interactions such as \textit{Link arms} and \textit{Massaging leg}, where strong physical contact and close-range coordination are essential.
Furthermore, even for interactions that involve minimal or no physical contact, like \textit{Imitate}, the DREM-enhanced model consistently achieves superior performance. This demonstrates that DREM effectively models dynamic interpersonal distance patterns, whether the interaction is contact-driven or contact-free.

We further analyze DREM in Tab. \ref{Table. Inter-X Distances Value} by examining several representative categories to gain deeper insights into its behavior. For each category, we compute the largest per-sequence displacement among all joint pairs between the two interacting individuals, and then average these values across all samples within the category to obtain an overall displacement measure $dsp$. Intuitively, a larger $dsp$ corresponds to more intense interactions characterized by substantial variations in mutual distance. As shown in Tab. \ref{Table. Inter-X Distances Value}, categories with larger relative displacement consistently exhibit greater performance improvements when DREM is incorporated. This explains why contact-heavy interactions such as  \textit{Massaging leg} benefit from a substantial AUC gain, whereas interactions like \textit{Fly kiss}, which involve minimal distance changes throughout the sequence, show limited improvement.

\noindent{\textbf{Effect of Parameter Sharing Design.}}
Our TASM is designed with parameter sharing between the two streams to more effectively capture synchronized temporal patterns during interaction. To examine its impact, we conduct an ablation study in Tab. \ref{Table. Inter-X parameter-share}, where we compare our shared-parameter design with a variant in which the two streams are learned independently. The results show that parameter sharing leads to notably higher detection performance, indicating that learning a unified temporal representation for both individuals is more beneficial than treating their trajectories in isolation. Without shared parameters, the model tends to over-specialize to person-specific motion patterns and struggles to generalize to relational interaction dynamics. We can thus verify the effectiveness of our parameter-sharing design in constituting the TASM in IADNet.

\begin{table}
\caption{\textbf{Ablation studies} on parameter sharing schemes in TASU with average AUC on Inter-X. The best AUC scores are denoted in bold.}
    \centering    \begin{tabular}{cc}
    \toprule
    Method& Avg.\\
    \hline
    Our w/o param-share& 0.624\\
    Our w param-share& \textbf{0.707}\\
    \bottomrule
    \end{tabular}
    \label{Table. Inter-X parameter-share}
\end{table}

\section{Conclusion}
In this paper, we propose the \textit{first} benchmark dedicated to human-human interaction detection (H2IAD), which aims to comprehensively examine anomalous interpersonal interactions. We further propose \textbf{IADNet}, a Transformer-based framework designed specifically for H2IAD. By formulating a pair of parameter-sharing encoding streams, IADNet explicitly models both the synchronous and asymmetric interpersonal dynamics, thereby producing collaboration-aware feature embeddings tailored for anomaly detection. To improve social understanding, we additionally incorporate a relational encoding module that injects distance-aware cues between interacting individuals into the encoding phase. The NF model is eventually leveraged to achieve likelihood-based anomaly scoring.  Extensive experiments on two large-scale human-human interaction datasets demonstrate that IADNet effectively identifies anomalies across previously unseen yet semantically related interaction patterns to address the challenges of H2IAD.

\noindent \textbf{Limitations and future directions:} Despite the effectiveness, our benchmark still has some limitations: (a) Real-world human interactions often involve rapid transitions between multiple action patterns, forming long-range interaction sequences that may contain time-varying anomalies. In contrast, our current setting assumes a uniform anomaly label across the entire sequence. Extending the framework to support temporally localized anomaly detection would be a valuable direction for the future; (b) Certain samples in existing datasets (e.g., NTU RGB+D 120) contain substantial noise, including inter-person penetration artifacts. Such issues can negatively impact model training, particularly for our DREM module, which relies on accurate distance-based cues. Curating or constructing higher-quality datasets to advance the study of H2IAD can also be an interesting future direction.


%





\ifCLASSOPTIONcaptionsoff
  \newpage
\fi



%

\bibliographystyle{abbrv}
\bibliography{reference}

%

\end{document}